\documentclass[preprint,12pt,a4paper,3p]{elsarticle}

\usepackage{amssymb}

\usepackage{lineno}

\usepackage{color}
\usepackage{tikz}
\usepackage[symbol]{footmisc}
\usepackage{subfig}
\usepackage{acronym}
\usepackage{url}

\newcommand{\customyinyang}{%
    \begin{tikzpicture}[scale=0.45]
      \draw[line width = 0.05ex,transform canvas={yshift=0.12ex}] (0,0) circle (1ex);
      \path[fill=black,transform canvas={yshift=0.12ex}] (90:1ex) arc (90:-90:0.5ex)
                        (0,0)    arc (90:270:0.5ex)
                        (0,-1ex) arc (-90:-270:1ex);
    \end{tikzpicture}}
    
\newcommand\blfootnote[1]{%
  \begingroup
  \renewcommand\thefootnote{}\footnote{#1}%
  \addtocounter{footnote}{-1}%
  \endgroup
}

\newcommand{\tabref}[1]{Table~\ref{#1}}
\newcommand{\figref}[1]{Figure~\ref{#1}}
\newcommand{\eqref}[1]{Equation~(\ref{#1})}

\usepackage{float}
\restylefloat{table}

\journal{SoftwareX}

\begin{document}

\begin{frontmatter}


\title{ROS-Mobile: An Android\textsuperscript{\texttrademark} application for the Robot Operating System}

\author{Nils Rottmann$^{1\customyinyang*}$, Nico Studt$^{1\customyinyang}$, Floris Ernst$^{1}$, Elmar Rueckert$^{1}$}

\address{$^{1}$Institute for Robotics and Cognitive Systems, University of Luebeck, 23562 Luebeck, Germany}

\begin{abstract}

\blfootnote{$^{\customyinyang}$These authors contributed equally to this work}\blfootnote{$^{*}$Corresponding Author: rottmann@rob.uni-luebeck.de}Controlling and monitoring complex autonomous and semi autonomous robotic systems is a challenging task. The Robot Operating System (ROS) was developed to act as a robotic middleware system running on Ubuntu Linux which allows, amongst others, hardware abstraction, message-passing between individual processes and package management. However, active support of ROS applications for mobile devices, such as smarthphones or tablets, are missing. We developed a \acs{ros} application for Android, which comes with an intuitive user interface for controlling and monitoring robotic systems. Our open source contribution can be used in a large variety of tasks and with many different kinds of robots. Moreover, it can easily be customized and new features added. In this paper, we give an outline over the software architecture, the main functionalities and show some possible use-cases on different mobile robotic systems. 

\end{abstract}


\end{frontmatter}
\section*{Metadata}
\vspace{-10pt}
\begin{table}[H]
\begin{tabular}{l p{6cm}}
\hline
Current code version & 1.0  \\
Permanent link to code/repository & \url{https://github.com/ROS-Mobile/ROS-Mobile-Android} \\
Legal Code License & The MIT License (MIT) \\
Code versioning system used & git \\
Software code languages, tools and services used & Java, XML, Gradle \\
Compilation Requirements and Dependencies & Linux or Windows with Android Studio Version 3.6.1 or higher \\
Developer Documentation & \url{https://github.com/ROS-Mobile/ROS-Mobile-Android/wiki} \\
User and developer support & rosmobile.info@gmail.com \\
\hline
\end{tabular}
\end{table}


\section{Introduction}
\begin{figure}
\subfloat[Different operating system for mobile devices \cite{MobileOp72:online}.]{\includegraphics[width=0.49\textwidth]
{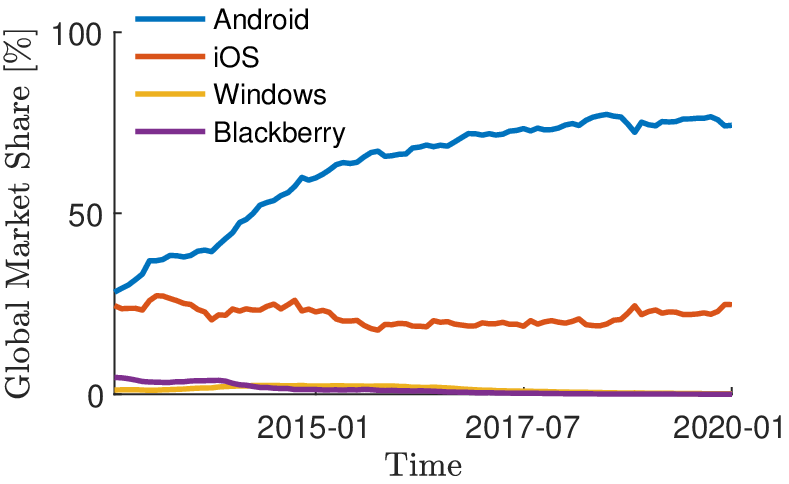}}
\hspace{0.01\textwidth}
\subfloat[Desktop, Mobile and Tablet devices \cite{Desktopv59:online}]{\includegraphics[width=0.49\textwidth]{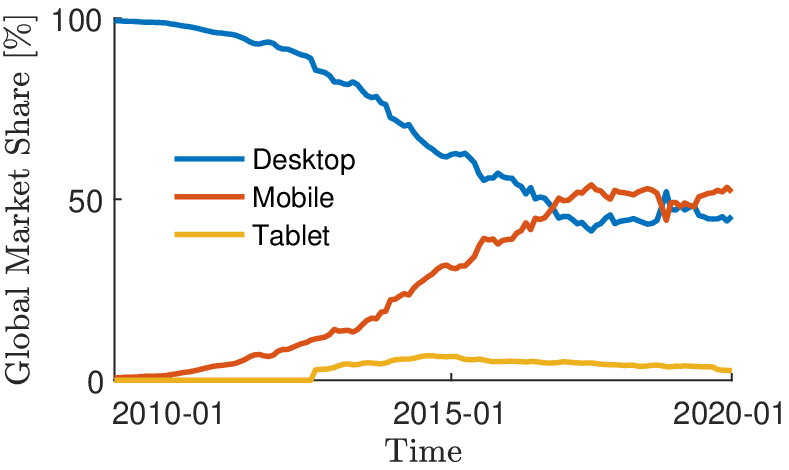}}
\caption{The global market share for (a) different operating systems and (b) different devices.}
\label{fig:MarketShare}
\end{figure}

The \ac{ros} \cite{quigley2009ros} is a widely used framework at universities and companies for control and navigation of robotic systems. For example for controlling \ac{uav} using \ac{mpc} methods \cite{kamel2017model} or mapping the environment with mobile robots using 2D \ac{slam} techniques \cite{santos2013evaluation}. For visualizing robot data, setting marker positions or boundary perimeter the \ac{rviz}  \cite{hershberger2008rviz} can be used which comes with a full-desktop installation. However, \ac{rviz} is not available for mobile devices running with Android on which $75 \, \%$ of all mobile devices operate, \figref{fig:MarketShare}a. Hence, to facilitate the interaction with different robotic systems, an Android application is required that on the one hand provides different control and visualization options similar to \ac{rviz} and on the other hand is adaptable and intuitive to use. Moreover, it should include additional functionalities not available in \ac{rviz} but provided by other \ac{ros} packages, e.g. teleoperational control \cite{rosteleo7:online}. Such an application allows the operator/user to efficiently control and supervise robots via smartphone or tablet which enables scientists to conduct easily outdoor experiments. All the more, since the share of mobile devices is steadily increasing which is shown in \figref{fig:MarketShare}b. For example, we use our application to control and monitor an autonomous lawn mower testing new mapping algorithms \cite{rottmann2019loop}, a mobile agent mapping and navigating through unknown terrain for guiding patients through, for example, a hospital \cite{MIRANAby83:online} or for education purposes.\\

\subsection{Related Work}

\begin{table}[]
    \centering
    \begin{small}
    \begin{tabular}{c|ccccc}
        Application & Android Version & Features & Customizable & MVVM & Support  \\
        \hline
        ROS Control & 4.0+ & Various & Yes & - & No \\
        AuTURBO & 4.0+ & Teleoperation & No & - & No \\
        \ac{ros} Image Viewer & 2.3+ & Image View & No & - & No \\
        \textbf{\ac{ros}-Mobile} & \textbf{5.0+} & \textbf{Various} & \textbf{Yes} & \textbf{+} & \textbf{Yes}
    \end{tabular}
    \caption{Comparison between different \ac{ros} applications currently available for Android.}
    \label{tab:comparison}
    \end{small}
\end{table}

There have been different attempts and studies regarding remote control and monitoring of robots via Android mobile devices. For example, in \cite{barbosa2015ros}, a low cost robot framework, controlled by two different Android applications based on ROSJava \cite{kohler2011rosjava}, was introduced. In \cite{nadvornik2014remote}, a remote control application for mobile robots built from Lego Mindstorms NXT kit was proposed. However, these applications are limited to specific robotic system (e.g. Lego Mindstorms NXT kit) or a specific robot configuration. Moreover, ROSJava is limited to older Android versions and programming architectures, especially with the nowadays recommended \ac{mvvm} architecture, which will be introduced later. Currently, there are different \ac{ros} applications for Android available for download, \tabref{tab:comparison}, which all suffer different short comings: \\
\textit{\ac{ros} Control} \cite{mtbiiRob1:online}, the most advanced application, allows for example map view, remote control or laser scan visualization. However, it is based on outdated Android versions and is not supported any longer. \textit{\ac{ros} Teleop Controller AuTURBO} \cite{AuTURBOr54:online} is mainly designed for control and display data supported by the robots from the TurtleBot family \cite{TurtleBo53:online}. \textit{\ac{ros} Image Viewer} only allows to display images received from a \ac{ros} framework.\\

\subsection{\ac{ros}-Mobile}
We propose a highly customizable \ac{ros} application for Android operated devices for remote control and to monitor mobile robotic systems. The application is based on the \ac{mvvm} architecture which comes with simple testing structures and extensibility. In order to use our application, only a mobile device, e.g. a smartphone or a tablet, with Android Version 5.0 (Lollipop) or higher is required. Additionally, a wireless ethernet connection between the \ac{ros} master and the mobile device has to be established. After starting our application, a new configuration can be set up or an already existing one can be loaded. A new configuration requires to set the correct \ac{ros} master IP. Required \ac{ros} nodes can then be added in the \textit{Details} section and customized. The remote control and the visualization can then be accessed in the \textit{Viz} section.\\

\section{Software description}
\label{se:description}
In the following, we describe the basic architecture, called \textit{\ac{mvvm}},  which our Android application is based on. This standard architecture allows the development of robust, efficient applications. In addition, we give an overview over the main functionalities of our application and demonstrate how they can be integrated in a common \ac{ros} workflow.  

\subsection{Software Architecture}
\label{se:architecture}

\begin{figure}
\centering
\includegraphics[width=0.49\textwidth]{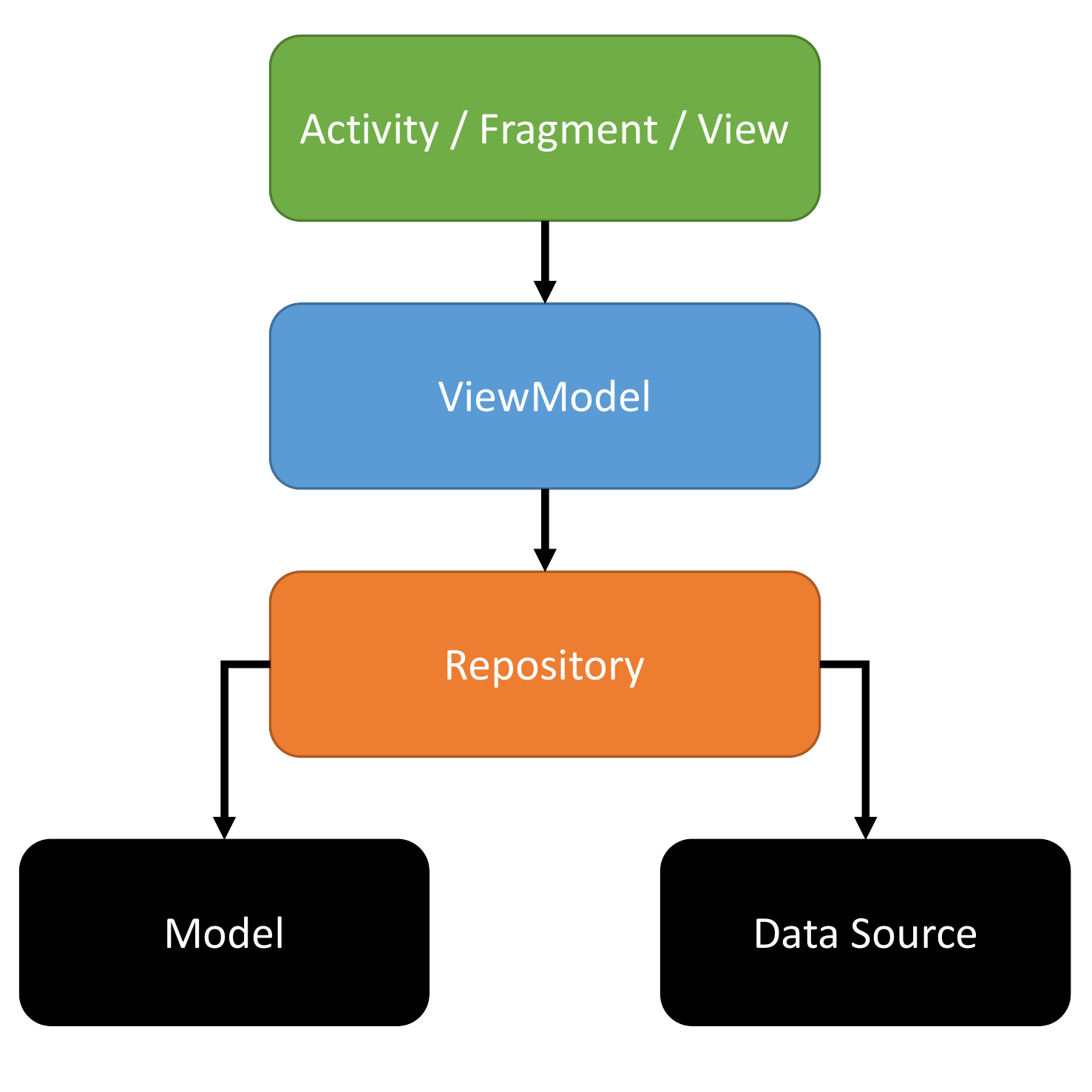}
\caption{The general \ac{mvvm} designing concept for mobile applications.}
\label{fig:mvvm}
\end{figure}

The basic architecture that our app is based on is \textit{\ac{mvvm}}, \figref{fig:mvvm}. The underlying principle is the \textit{Separation of Concerns} \cite{hursch1995separation}. This means that each \textit{concern} should be addressed by a different \textit{section} of our program. For example, we require a \ac{ui}, to enable user interaction, as well as \ac{ros} nodes, to subscribe or publish to certain \ac{ros} topics. The information produced by the user and the data received or sent via the \ac{ros} nodes are thereby the \textit{concerns} of our program which have to be handled separately. Moreover, Android applications in comparison to desktop applications in general do not have a single entry point instead they are consisting of multiple components, such as Activities or Fragments. These application components are then used by the Android \ac{os} for integration into the device workflow. Hence, our application has to be able to handle app hopping behavior correctly. To do so, we use the \textit{ViewModel} class, which has been designed for managing \ac{ui} related data, as a connection between our model/data storage system and our \ac{ui}. The required data is provided by a model/data storage system and is requested by the \textit{ViewModel} object. This \textit{ViewModel} object is directly assigned to a \ac{ui} Controller (e.g. Activity, Fragment), which requests the data from the \textit{ViewModel} object and let the user interact with them. Thus, the \ac{ui} controller is decoupled from the data management system and can be destroyed and rebuilt during runtime by the \ac{os} to enhance performance. In order to implement such type of architecture, the ROSJava code, which was designed for older Android versions and architectures had to be modified according to the requirements of the \ac{mvvm} architecture.

\subsection{Software Functionalities}
\label{se:functionalities}
Our \ac{ros} Android application enables Android devices to easily control \ac{ros} operated systems. Therefore, the application establishes a \ac{ros} connection via WLAN with the system. This allows to control the robot by publishing \ac{ros} messages (e.g. velocity commands), visualize published measurement results (e.g. map data) or analyze the system's behavior. Moreover, an additional control possibility is given by the included SSH client, which allows direct access to the operating system. All in all, \ac{ros} operated systems can be easily manipulated using the proposed \ac{ros} Android application which makes it much easier to work with mobile robots outside laboratory environments.

\section{Application Example}
\label{se:Example}

\begin{figure}
\subfloat[Master Connection\label{fig:MasterLayout}]{\includegraphics[width=0.23\textwidth]{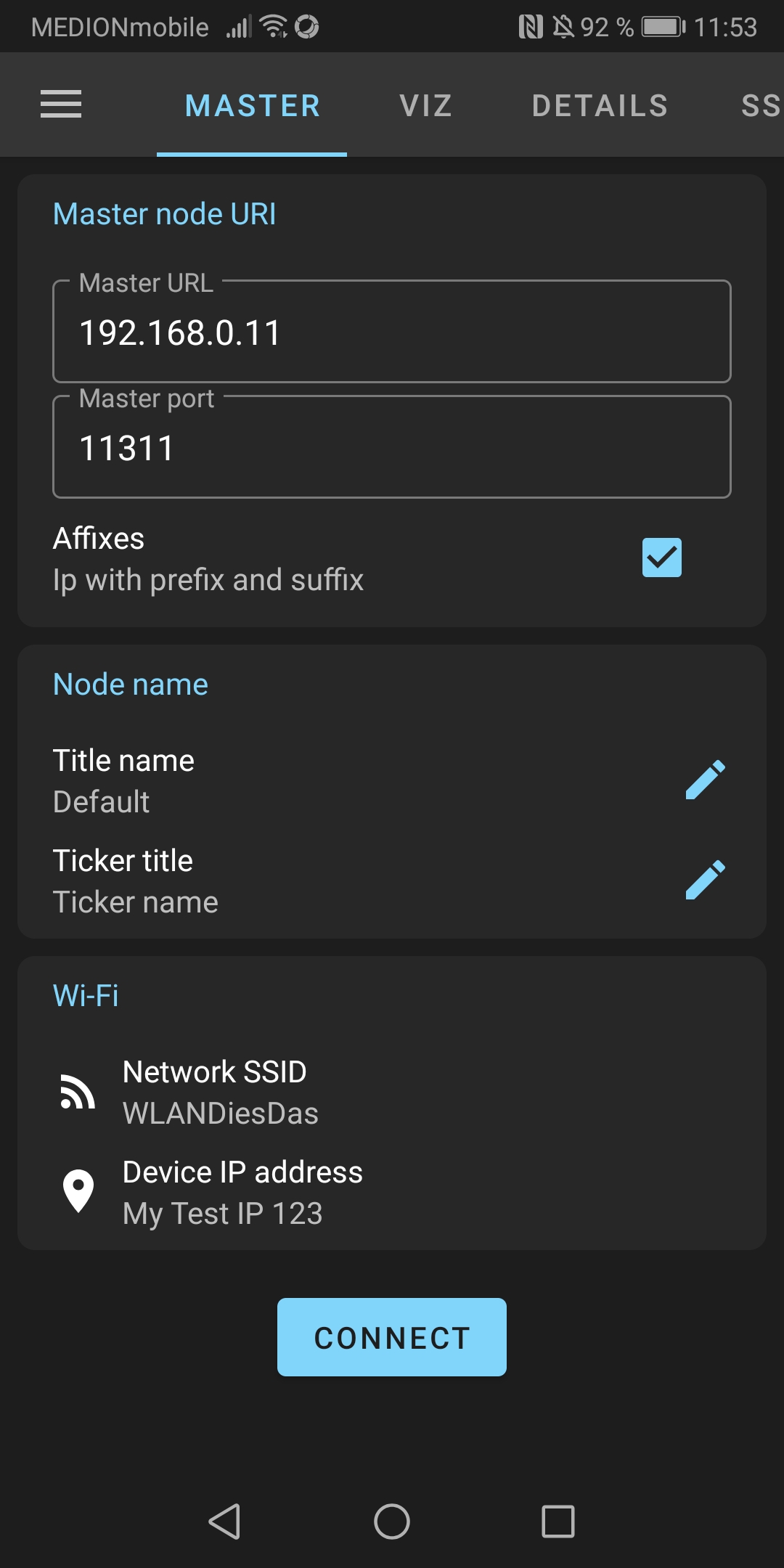}}
\hspace{0.01\textwidth}
\subfloat[Visualization Details\label{fig:DetailsLayout}]{\includegraphics[width=0.23\textwidth]{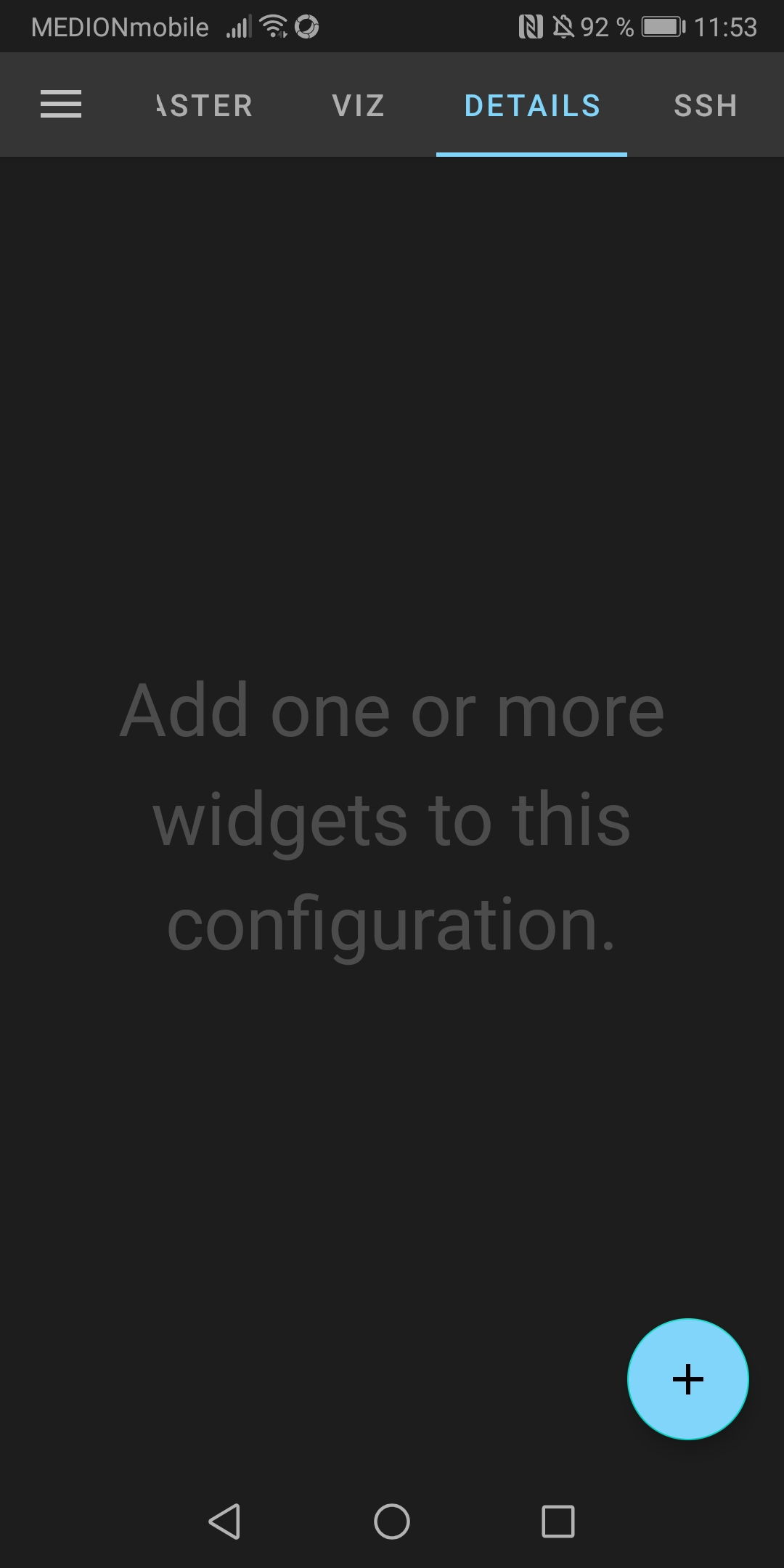}}
\hspace{0.01\textwidth}
\subfloat[Added Nodes\label{fig:Nodes}]{\includegraphics[width=0.23\textwidth]{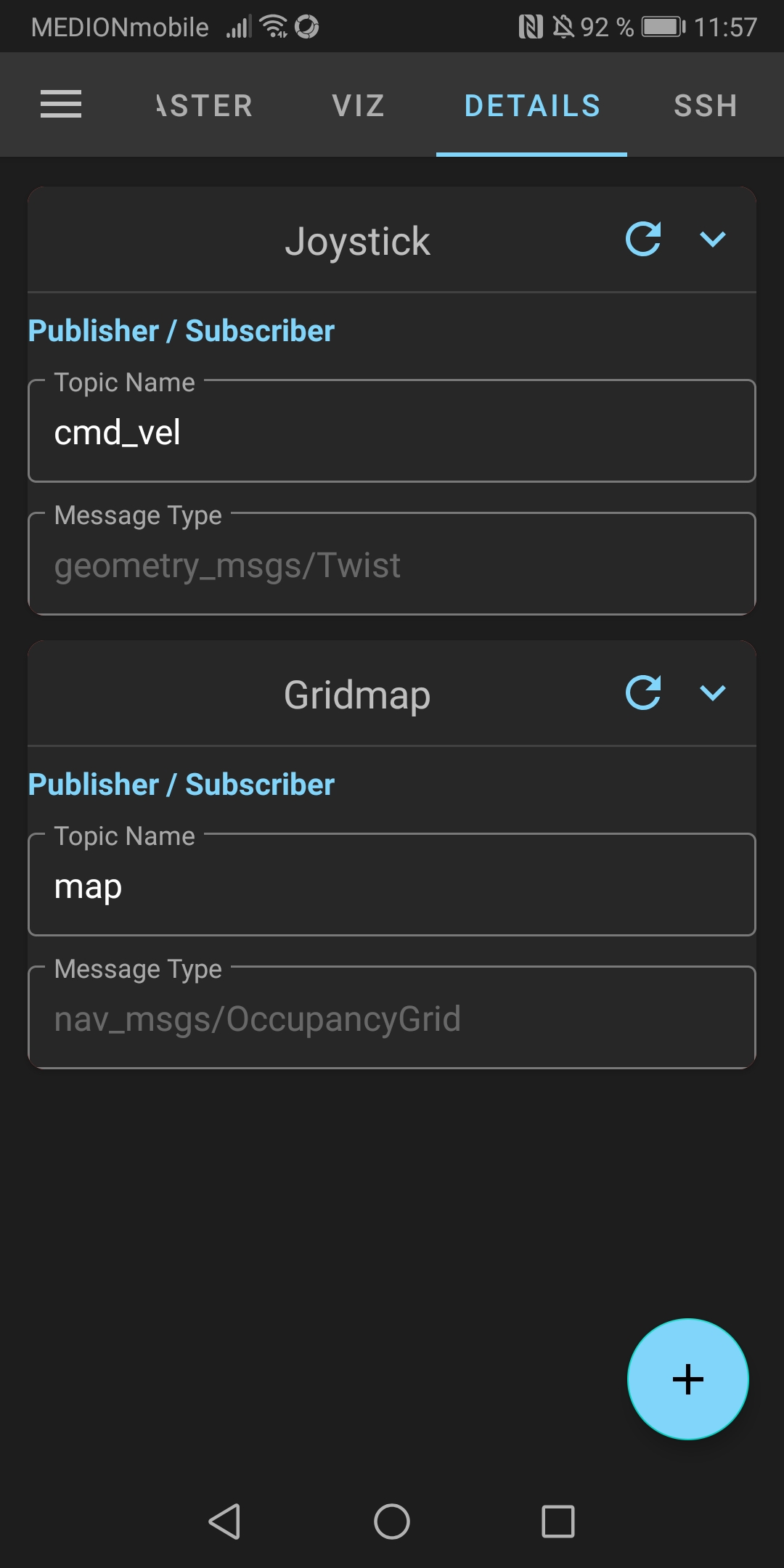}}
\hspace{0.01\textwidth}
\subfloat[Visualization\label{fig:VizLayout}]{\includegraphics[width=0.23\textwidth]{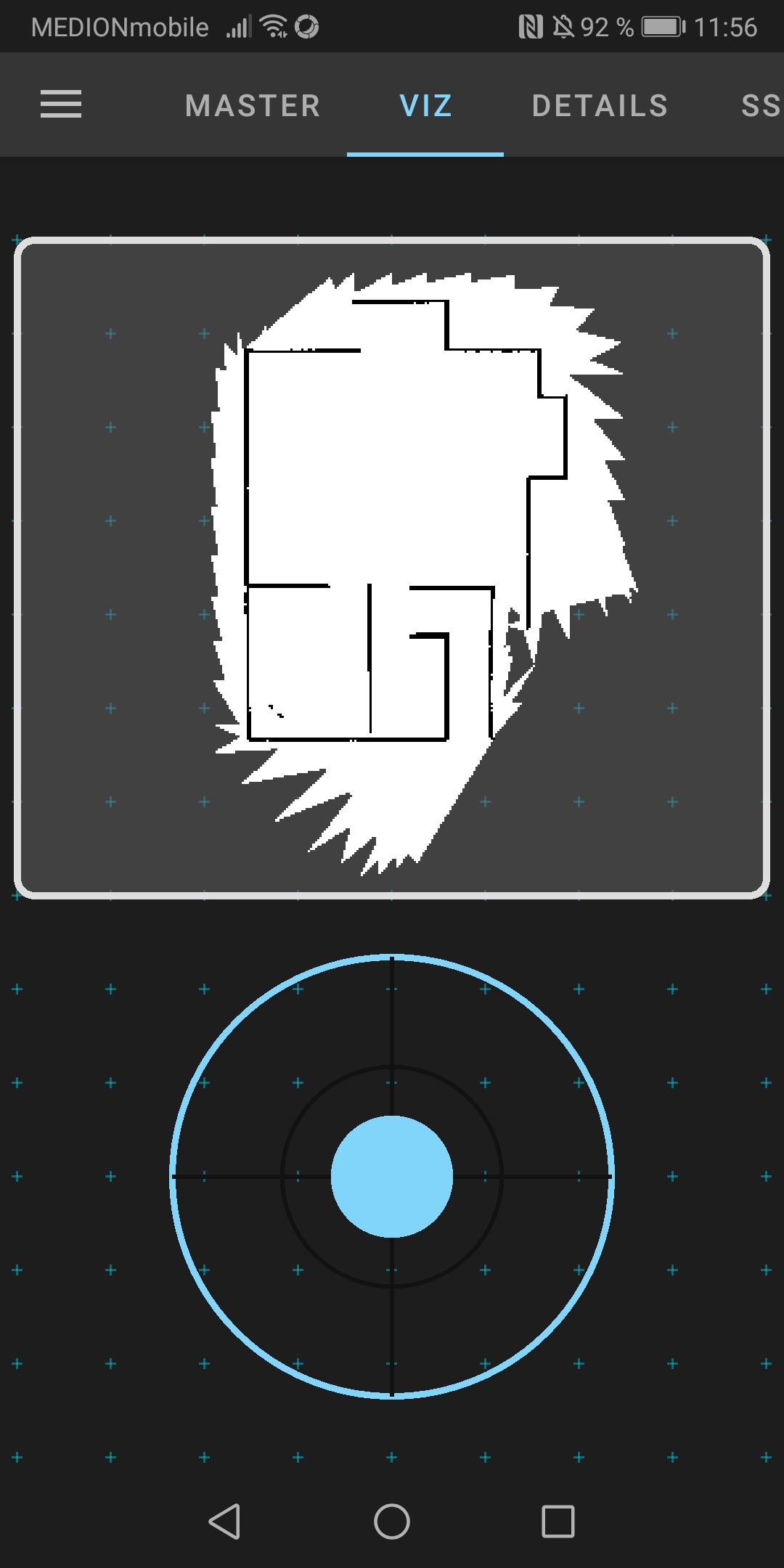}}
\caption{Illustrative example for the mapping of an apartment environment and displaying the mapping result with \ac{ros}-Mobile.}
\label{fig:illuExample}
\end{figure}

The goal of this task was to build a map of an apartment environment using a differential drive robot. Therefore, we connected the application with the \ac{ros} master running the differential drive robot over WLAN by inserting the correct IP address in the \textit{MASTER} configuration tab, \figref{fig:MasterLayout}. Adding \ac{ros} nodes in the details tab, \figref{fig:DetailsLayout} and \figref{fig:Nodes}, enables the control of the differential drive robot via a joystick method sending \textit{geometry\_msgs/Twist} to a \textit{cmd\_vel} topic and the visualization of the generated occupancy grid map by subscribing to the \textit{map} topic via a gridmap method. In the visualization tab, \figref{fig:VizLayout}, the recorded occupancy grid map is displayed as well as the joystick. The joystick can be used via touch for sending control inputs over the \textit{cmd\_vel} topic to the differential drive robot. For a more detailed example, we refer to the explanatory video\footnote{https://www.youtube.com/watch?v=T0HrEcO-0x0}.

\section{Current Application Domains}
\label{se:impact}

\ac{ros}-Mobile is an Android application which allows direct control and data visualization for mobile robots using \ac{ros}. It can be used to easily conduct or supervise outdoor experiments with mobile robots using mobile devices, such as smartphones or tablets. Thus, it enables researchers to simplify field experiments and hence, accelerates the evaluation time for mobile robotic devices. Currently, we use the \ac{ros}-Mobile in different research projects, \figref{fig:MobileRoboticPlatforms}, which we shortly summarize in the following. 

\subsection{Household Robots}
\begin{figure}
\subfloat[Intelligent Lawn Mower\label{fig:LawnMower}]{\includegraphics[width=0.31\textwidth]{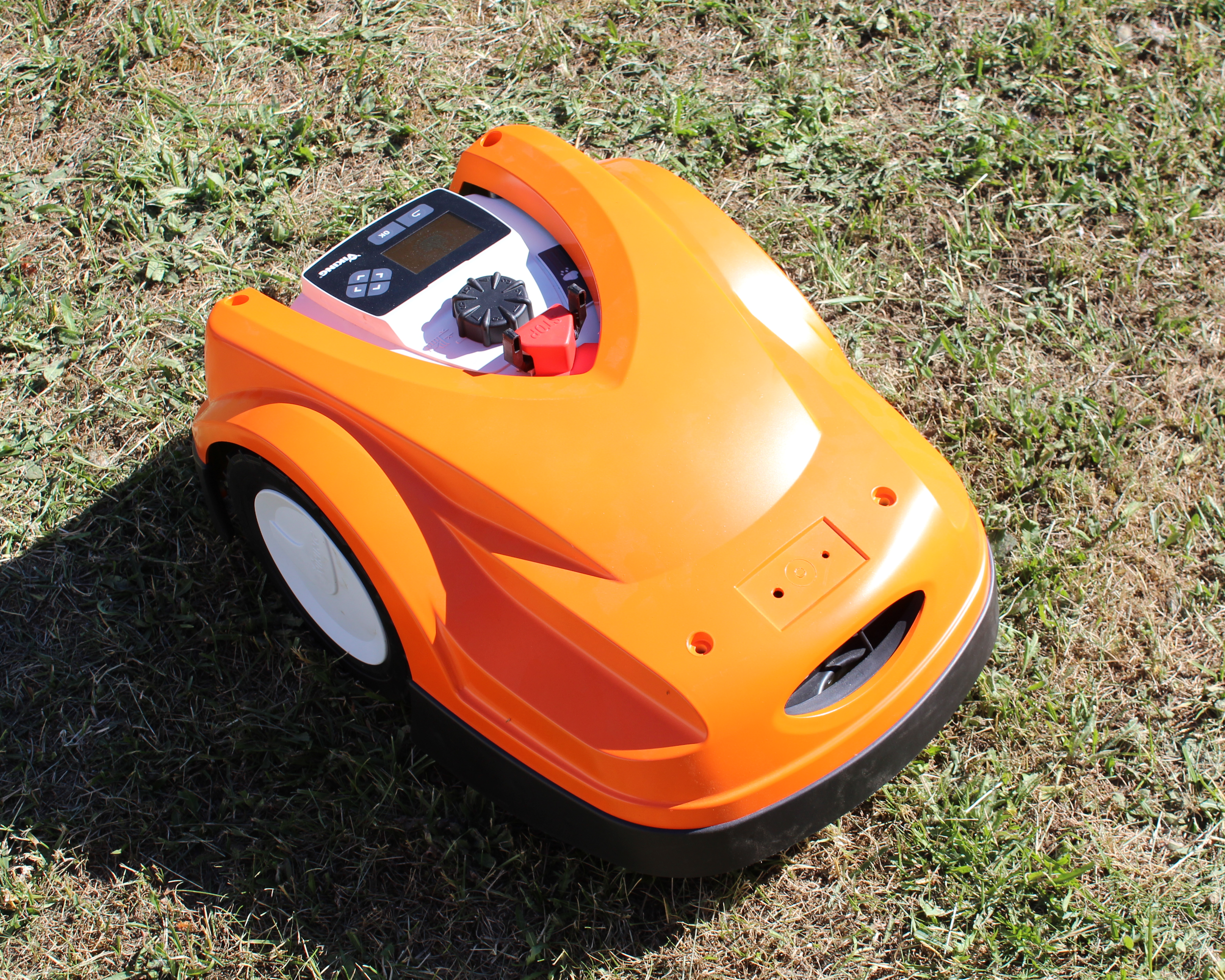}}
\hspace{0.01\textwidth}
\subfloat[Autonomous Surface Vehicle\label{fig:ASV}]{\includegraphics[width=0.31\textwidth]{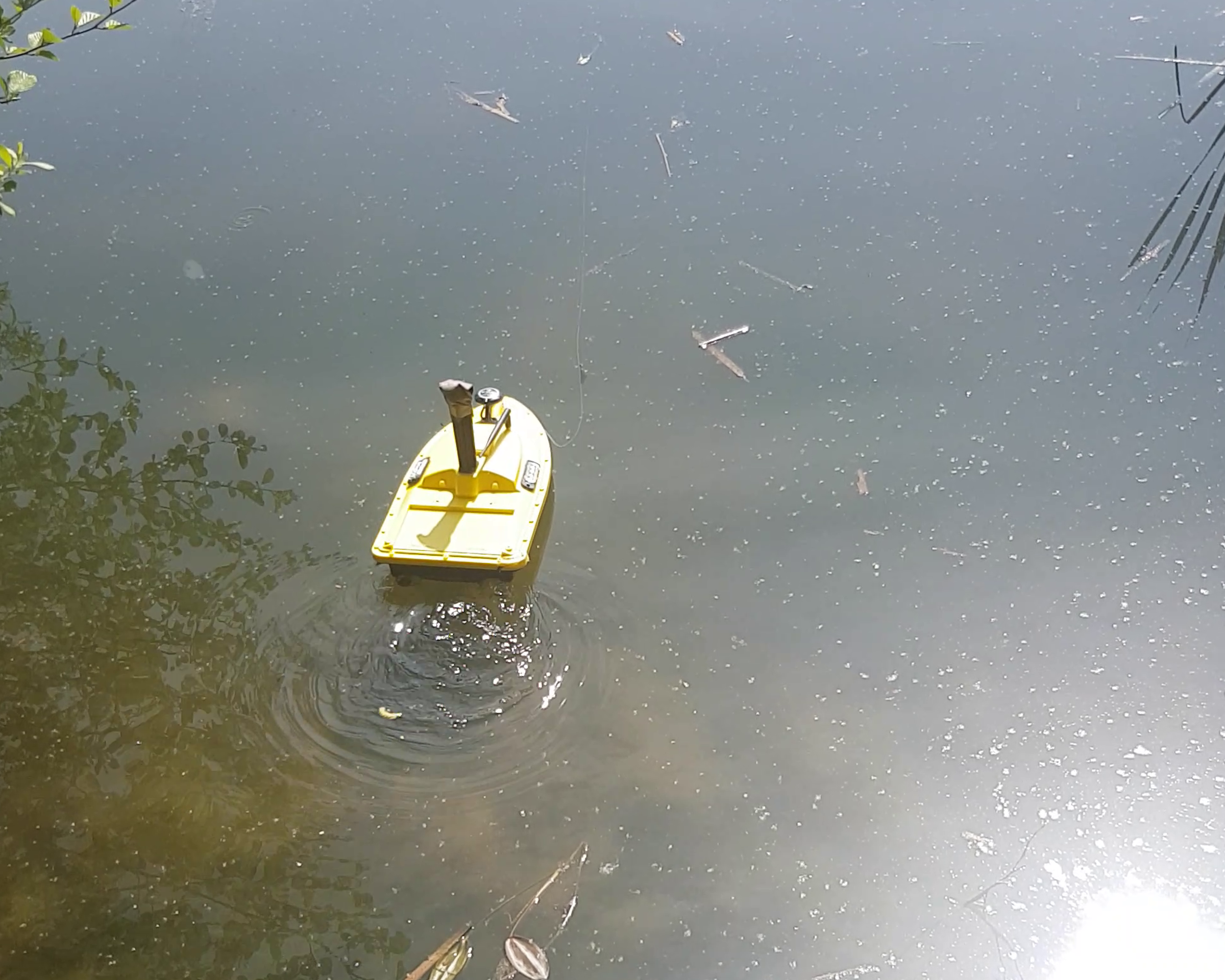}}
\hspace{0.01\textwidth}
\subfloat[Loomo, Segway Robot\label{fig:Loomo}]{\includegraphics[width=0.31\textwidth]{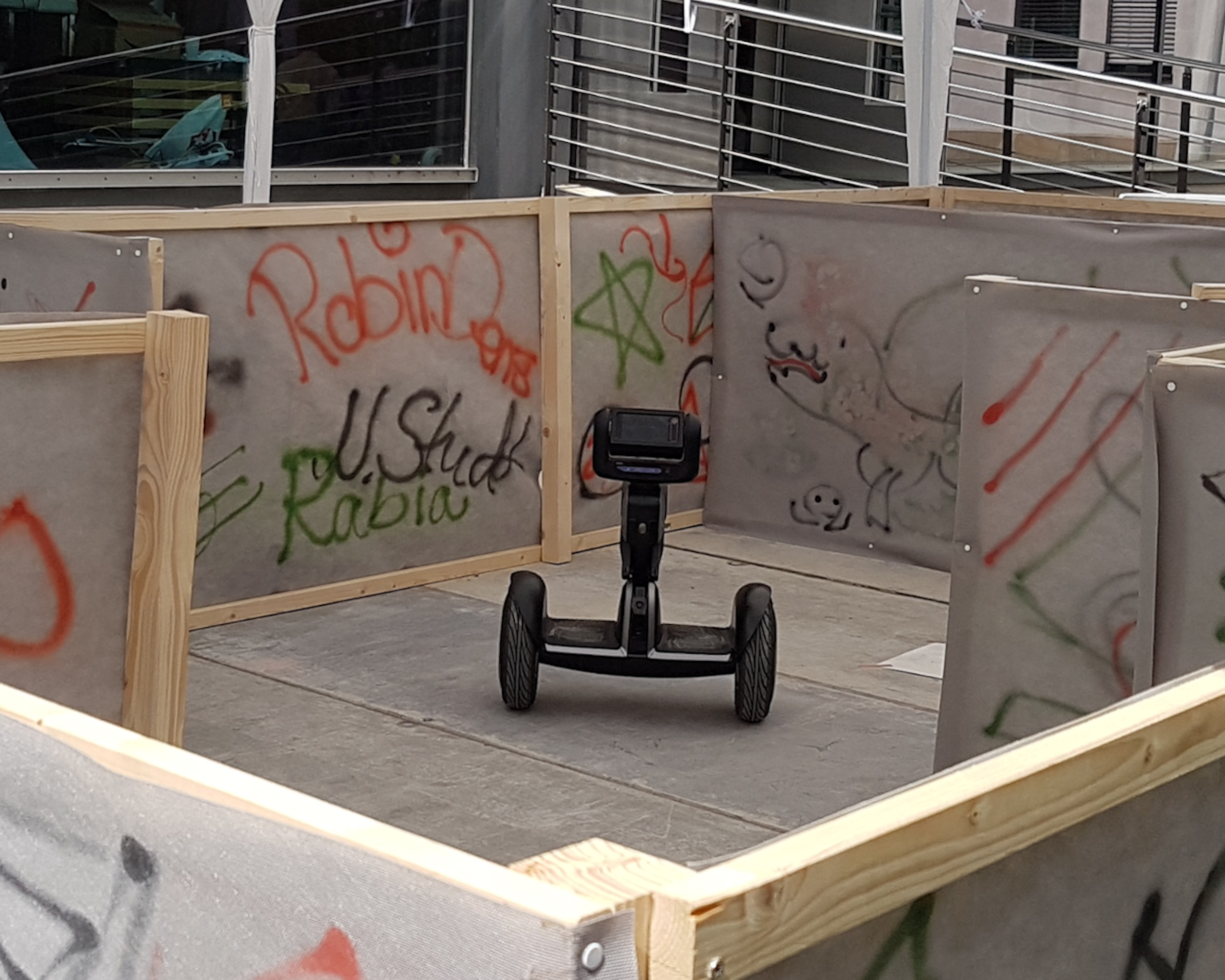}}
\caption{Mobile robotic platforms for which \ac{ros}-Mobile is currently used for evaluation and control.}
\label{fig:MobileRoboticPlatforms}
\end{figure}

 An area of increasing importance in the last decade is the field of low-cost robotics. Robots such as lawn mowers, \figref{fig:LawnMower}, or vacuum cleaners are used ubiquitously in households and work exclusively in closed environments, e.g. on a lawn or in an apartment. We develop low cost sensors, mapping and path planning algorithms for such robots, which are tested in real garden scenarios using the \ac{ros}-Mobile application. Among other things, the joystick node and the SSH connection have been used intensively. 

\subsection{\ac{adrz}}
The \ac{adrz} project investigates the use of robot systems for civil terrestrial danger prevention in inhospitable environments. As part of the \ac{adrz}, we explore the use of intelligent \ac{asv} and quadrocopters for fast detection of hazardous substances and flooding. For testing these robotic exploration approaches in realistic outdoor environments, the \ac{ros}-Mobile application is used for remote control and visualization.

\subsection{\ac{mirana}}
Intelligent navigation and information services will make the visit in hospitals a better experience. In the \ac{mirana} project, we build the software that enables robots to guide humans autonomously in the healthcare environment. We are starting with the software solution for \ac{slam} in hospital buildings, speaker tracking and speech recognition and synthesis. To that end, we are using the personal transporter robot from Segway Robotics and developing applications on the Segway Robot platform. For visualization during the testing of the different approaches, we use the \ac{ros}-Mobile application.

\section{Conclusions}
\label{se:conclusion}

We introduced \ac{ros}-Mobile, an android application for the robot operating system. We showed that \ac{ros}-Mobile allows fast and efficient control and supervision for realistic experiments with mobile robotic systems in outdoor as well as in indoor environments. Thus, it accelerates the development process for new sensors, algorithms or methods. The current version of \ac{ros}-Mobile implements different nodes, e.g. for teleoperational control or grid map visualization. Additional nodes and visualization methods can be added easily by following the tutorial on the GitHub repository. Contributions are welcome since \ac{ros}-Mobile is constantly being developed.

\section{Conflict of Interest}

We wish to confirm that there are no known conflicts of interest associated with this publication and there has been no significant financial support for this work that could have influenced its outcome.

\section*{Acknowledgements}
\label{se:Ack}

This project has received funding from the Deutsche Forschungsgemeinschaft (DFG, German Research
Foundation) No \#430054590 (TRAIN) and from the Bundesministerium für Bildung und Forschung (BMBF, Federal Ministry of Education and Research) No \#13N14862 (A-DRZ). We also like to thank Robin Denz and Ralf Bruder for their support by producing the introduction video.

\section*{Abbreviations}
\begin{acronym}[ECU]
\acro{ros}[ROS]{Robot Operating System}
\acro{uav}[UAV]{Unmanned Aerial Vehicle}
\acro{mpc}[MPC]{Model Predictive Control}
\acro{slam}[SLAM]{Simultaneous Localization and Mapping}
\acro{rviz}[rviz]{ROS Visualization}
\acro{mvvm}[MVVM]{Model View ViewModel}
\acro{ui}[UI]{User Interface}
\acro{os}[OS]{Operating System}
\acro{asv}[ASV]{Autonomous Surface Vehicles}
\acro{adrz}[A-DRZ]{Aufbau des Deutschen Rettungsrobotik-Zentrums}
\acro{mirana}[MIRANA]{Mobile Intelligent Robotic Agent for Navigation and Assistance}
\end{acronym}







\bibliographystyle{IEEEtran}
{\small
\bibliography{Paper}}




\end{document}